\documentclass[10pt,twocolumn,letterpaper]{article}

\usepackage[accsupp]{axessibility}
\usepackage{wacv}              

\usepackage{graphicx}
\usepackage{amsmath}
\usepackage{amssymb}
\usepackage{booktabs}

\usepackage{multirow}
\usepackage{multicol}
\usepackage[ruled,vlined]{algorithm2e}

\usepackage{color}
\usepackage[dvipsnames]{xcolor}
\usepackage{subcaption}
\newcommand{\blue}[1]{{\color{black}#1}}

\newcommand{\bW}{\mathbf{W}}

\newcommand{\calD}{\mathcal{D}}

\newcommand{\calL}{\mathcal{L}}

\newcommand{\calI}{\mathcal{I}}
\newcommand{\calF}{\mathcal{F}}

\newcommand{\calX}{\mathcal{X}}

\usepackage[pagebackref,breaklinks,colorlinks]{hyperref}

\usepackage[capitalize]{cleveref}
\crefname{section}{Sec.}{Secs.}
\Crefname{section}{Section}{Sections}
\Crefname{table}{Table}{Tables}
\crefname{table}{Tab.}{Tabs.}

\def\wacvPaperID{1238} 
\def\confName{WACV}
\def\confYear{2024}

\begin{document}

\title{Frequency Attention for Knowledge Distillation}

\author{%
Cuong Pham\textsuperscript{\rm 1}
\and
Van-Anh Nguyen\textsuperscript{\rm 1}
\and
Trung Le\textsuperscript{\rm 1}
\and
Dinh Phung\textsuperscript{\rm 1}
\and
Gustavo Carneiro\textsuperscript{\rm 2} ~~~~~~~~~
Thanh-Toan Do\textsuperscript{\rm 1}
\\
\textsuperscript{\rm 1}Department of Data Science and AI, Monash University, Australia \\
\textsuperscript{\rm 2}Centre for Vision, Speech and Signal Processing, University of Surrey, United Kingdom 
}

\maketitle


\begin{abstract}
Knowledge distillation is an attractive approach for learning compact deep neural networks,
 which learns a lightweight student model by distilling knowledge from a complex teacher model. Attention-based knowledge distillation is a specific form of intermediate feature-based knowledge distillation that uses attention mechanisms to encourage the student to better mimic the teacher. However, most of the previous attention-based distillation approaches perform attention in the spatial domain, which primarily affects local regions in the input image. This may not be sufficient when we need to capture the broader context or global information necessary for effective knowledge transfer. In frequency domain, since each frequency is determined from all pixels of the image in spatial domain, it can contain global information about the image.
Inspired by the benefits of the frequency domain, we propose a novel module that functions as an attention mechanism in the frequency domain. The module consists of a learnable global filter that can adjust the frequencies of student's features under the guidance of the teacher's features, which encourages the student's features to have patterns similar to the teacher’s features. We then propose an enhanced knowledge review-based distillation model by leveraging the proposed frequency attention module. The extensive experiments with various teacher and student architectures on image classification and object detection benchmark datasets show that the proposed approach outperforms 
\blue{other} knowledge distillation methods. 
\end{abstract}

\section{Introduction}
\label{sec:intro}
Convolutional Neural Networks (CNNs) have been widely applied and achieved a myriad of successes in various computer vision tasks, such as image classification, object detection, and image segmentation. However,  {these CNNs} often  {require} expensive memory and computation resources, 
{making them} unsuitable for applications with limited resources. Different approaches have been proposed to learn efficient deep neural networks, such as pruning \cite{prunning,channel_pruning,ThiNet}, knowledge distillation \cite{KD,FitNets, OFD, ReviewKD}, and quantization \cite{QuantizedCNN,LQNets,dorefa, pham2023collaborative}. Among them, knowledge distillation (KD) is an attractive approach to reduce the computational cost of CNNs.   
In knowledge distillation, a smaller student network is trained to mimic the behaviour of a larger teacher network. 

Different approaches have been proposed for knowledge distillation~\cite{KD, FitNets, AT, CRD, WCoRD, OFD, ReviewKD,DKD}. Among them, intermediate feature-based KD is a popular approach because it is flexible to design different distillation mechanisms such as layer to layer distillation~\cite{FitNets,AT,OFD} and layer fusion distillation~\cite{ReviewKD}. 
Attention-based KD \cite{AT, AFD, ASKD, DenseAT} is a specific form of intermediate feature-based knowledge distillation. In those works, the attention is performed in the spatial domain and they use attention maps to help the student to focus on the most informative information from the teacher. However, in ~\cite{AT, AFD, ASKD, DenseAT} each value of the attention map is calculated from a local region of the input feature map. This focus of the local regions may not be sufficient to effectively transfer knowledge from teacher model to student model in knowledge distillation when we need to capture the broader context or global information necessary for effective knowledge transfer.

Our goal is to encourage student model to capture both detailed and higher-level information such as object parts from the teacher model. 
This can be accomplished by processing the student's features in the frequency domain instead of the spatial domain. The frequency domain is useful for understanding images with repetitive or periodic patterns that may be difficult to discover using traditional spatial domain techniques.  
\blue{By capturing the intensity changes and patterns in the image, the frequency domain can identify different regions associated with objects, and each frequency could correspond to some specific structures, e.g., high frequencies correspond to large changes in image intensity over a short pixel distance (e.g., edges).}

With the above benefits of the frequency domain, we propose a Frequency Attention Module (FAM), which has a learnable global filter in the frequency domain. The global filter can be seen as a form of attention in the frequency domain, which can adjust the frequency of student's feature maps. We then invert the attending features in frequency domain 
back to the spatial domain and minimize them with the teacher's features. By updating the parameters of the learnable filter based on the guidance of the teacher, we can encourage the transformed student's features to have similar patterns as the teacher's features. 

Given the proposed frequency-based attention module, we propose two enhanced architectures for layer-to-layer~\cite{OFD, AT, FitNets} and knowledge review distillation~\cite{ReviewKD}. We extensively demonstrate the effectiveness of our proposed method with various teacher and student architectures on benchmark datasets for image classification and object detection. The experimental results show that the proposed approach outperforms \blue{other} knowledge distillation methods. In summary, our contributions are:
\begin{itemize}   
    \item {We propose a novel module, which is our main contribution in which we explore Fourier frequency domain for knowledge distillation.  
    The module consists of a learnable global filter that can adjust frequency of the student's features, which encourages student's features to mimic patterns from the teacher’s features.}
    {\item We propose an enhanced layer-to-layer knowledge distillation model and an enhanced knowledge review-based distillation model by leveraging the proposed FAM module.} 
    {\item Our method outperforms \blue{other} knowledge distillation methods for classification on CIFAR-100 and ImageNet datasets and object detection on MS COCO dataset.}
\end{itemize}

\section{Related work}
\label{sec:related_work}

\textbf{Knowledge Distillation (KD)} 
has received substantial attention recently due to its versatility in various applications. In KD, the student model can benefit from the guidance of various forms from the teacher model to achieve better performance. This could be soft logit-based distillation \cite{KD, DKD}, relation-based distillation \cite{CRD, RKD, WCoRD}, or intermediate feature-based distillation~\cite{FitNets,AT, OFD, ReviewKD}. Among them, the feature-based knowledge distillation allows flexibility in designing distillation mechanisms. 
Particularly, in FitNets~\cite{FitNets} given a student layer (guided layer) and a teacher layer (hint layer), the authors minimize the $L_2$ distance between the transformed student's features and the teacher's features. 
Following FitNets, AT \cite{AT}, PKT \cite{PKT}, and SP \cite{SP} transfer knowledge through activation maps, feature distributions, and pairwise similarities, respectively. 
In OFD~\cite{OFD}, the authors propose margin ReLU applied on teacher's feature maps to select information used for distillation.      
In \cite{ReviewKD}, the authors introduce the review mechanism to enrich student features. They show that lower-level features of teacher are useful in supervising the higher-level features of student. They propose to fuse different levels of student features before mimicking teacher knowledge. 

In~\cite{AT, AFD, ASKD, DenseAT}, the attention is performed in the spatial domain and they use attention maps to help the student focus on the most informative information from the teacher.
Specifically, spatial attention maps in AT~\cite{AT} can be computed using the sum of absolute values across the channel dimension. AFD~\cite{AFD} also transfers knowledge from teacher to student through spatial attention maps that are computed through channel-wise
average pooling layer. They then maximize the similarity between attention maps of student's features and the attention maps of teacher's features. Meanwhile, ~\cite{ASKD} computes the spatial attention maps using average pooling and fully connected layers. 
However, with the attention in the spatial domain used~\cite{AT, AFD, ASKD, DenseAT}, weights of the attention map are usually calculated from local regions of the feature maps. The attention weights (i.e., values in the attention map) indicate the importance of the corresponding local regions. Due to its local property, a change in a value of the attention map (in backpropagation) only affects the corresponding local region. 

\vspace{1em}
\textbf{Fourier frequency domain and attention in the frequency domain.} 
In digital image processing, Fourier frequency domain represents an image with a set of sinusoidal waves, with each wave {representing} a different level of intensity in the whole image. The frequency domain is a helpful way to understand images that have repetitive or periodic patterns \cite{DIP}. It is more effective than traditional spatial domain techniques in capturing geometric structures that are difficult to extract. By capturing the intensity changes in the image, the frequency domain can identify distinct regions that are associated with objects. 

Each frequency in the frequency domain is determined by all the pixels in the image in the spatial domain. Frequencies can correspond to particular structures in the spatial domain. For instance, high frequencies correspond to significant changes in image intensity over a small distance between pixels, such as edges. \blue{Therefore, focusing on the frequency domain can be seen as a form of global attention.} Meanwhile, attention in the spatial domain \cite{AT, AFD, ASKD} primarily affects local regions in the input feature map, which may be insufficient for capturing the global structure of the feature map. 
\blue{By contrast, attention in the frequency domain can be especially useful for identifying global information or geometric structures of the feature map that may be difficult to detect using traditional spatial domain techniques. 
A change in a frequency of the attention frequency map
can impact the entire input feature,
compared to the effect on the local regions when changing a value in attention map in the spatial domain.} 


In this work, we explore the Fourier frequency domain for the knowledge distillation 
problem. We propose a frequency attention module (FAM) that has a learnable global filter, which acts as an attention in the frequency domain. Based on the guidance from the teacher, FAM will encourage the student's features to have similar patterns as teacher's features.

\section{Proposed method}
\label{sec:proposed}
This section first details the frequency attention module (FAM) that encourages the student to better mimic the teacher. We then present our design to integrate the FAM module into two popular knowledge distillation mechanisms, i.e., layer-to-layer feature-based distillation \cite{OFD} and knowledge review-based distillation \cite{ReviewKD}. 

\subsection{Frequency attention module} 
\label{subsec:FAM_module}
\begin{figure*}[]
\centering
  \includegraphics[scale=1.2, width=0.8\linewidth]{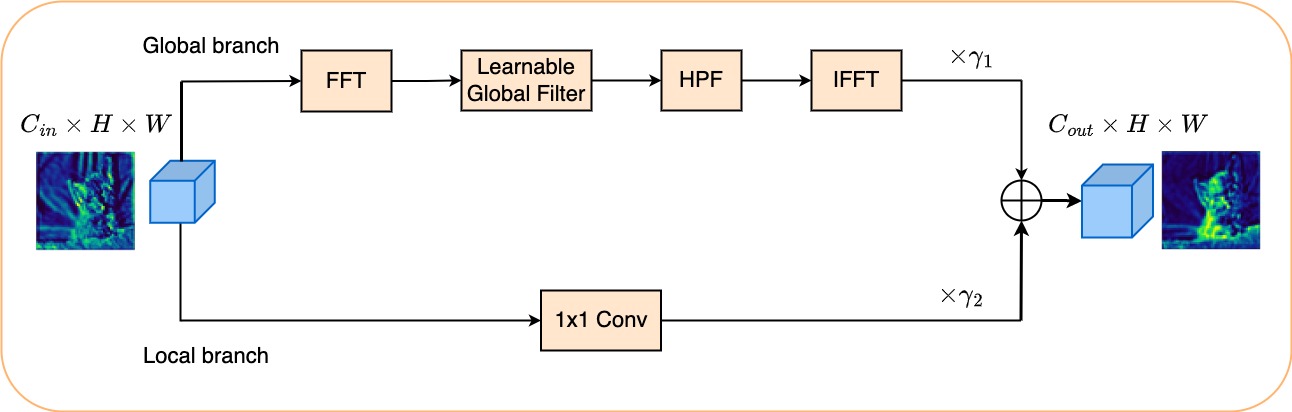}
    \caption{Fourier Frequency Attention Module. 
    HPF stands for a high pass filter. In the global branch, the input student's feature map is transformed to the frequency domain using the FFT. The frequency is then adjusted by a learnable global filter. A high pass filter is then applied to the adjusted frequency map to filter out lowest frequencies. The local branch consists of a 1$\times$1 convolutional layer in the spatial domain. The outputs of the global and local branches are added and the resulting feature map is compared with the teacher's feature map. $\gamma_1$ and $\gamma_2$ are the learnable weighting parameters of the global and local branches, respectively.}
\label{fig:FAM}       
\end{figure*}

As shown in Figure~\ref{fig:FAM}, the FAM module consists of global and local branches.
Specifically, given a feature map $X$ with a dimension of $C_{in} \times H \times W$, in the global branch we first transform it into the frequency domain via Fast Fourier Transform (FFT).
Here the FFT is applied to each channel separately. For the $i^{th}$ channel $X_i$ of the feature map $X$, the 2-D discrete FFT of $X_i$ denoted by $\calX_i$ is expressed as:
\begin{equation} \label{eq:FFT}
    \mathcal{X}_i(u, v) = \sum_{k=0}^{H-1}\sum_{l=0}^{W-1} X_i(k, l)e^{- {i2\pi} (\frac{uk}{H} + \frac{vl}{W})}.
\end{equation}

To adjust the frequencies of $\calX_i$, we apply a learnable global filter $K$ which can be seen as a form of attention on $\calX_i$.  
\paragraph{Global filtering.}
It is worth noting that in feature distillation, we want the feature map resulting from the FAM module to have the same dimension as the dimension of a given teacher feature map where the knowledge will be distilled. Therefore, we design the global filter $K$ with the dimension of $C_{out} \times C_{in} \times H \times W$, where $C_{out}$ is the number of channels of the teacher's feature map. Each kernel in the
global filter  $K$ has the same size as the 3D input tensor $\mathcal{X}$ with the size $C_{in} \times H \times W$. This kernel performs element-wise multiplied with the 3D input tensor $\mathcal{X}$, resulting in a 3D feature map with the same size as the input feature map.
 Next, the 3D frequency feature maps of the output are then summed up, (i.e., sum-pooling in each $C_{in} \times 1\times 1$ block), resulting in a 2D output with the size $H \times W$. The above operation is performed for $C_{out}$ kernels of the global filter, resulting in a 3D feature map with a size of $C_{out} \times H \times W$ as the output.
 
 It is worth noting that the proposed filter acts in the frequency domain.
 Each frequency in the frequency domain is determined by all the pixels in the spatial domain; 
 hence, although each element of each kernel attends to a particular frequency, the filter still achieves the global effects.

After that, we further suppress low frequencies, which encourages the student to {de-focus from} the non-salient regions. To this end, we add a high pass filter (HPF) after the learnable global filter to eliminate part of the lowest frequency components. The HPF is applied to each channel separately. Specifically, for each channel, we adopt the ideal HPF, which suppresses 1 percent of the lowest frequencies.

We then transform the frequency domain back to the spatial domain via the inverse Fast Fourier Transform (IFFT). Given $\Bar{\calX}$ which is the frequency feature map after the HPF, for the $i^{th}$ channel $\Bar{\calX_i}$ of the frequency feature map $\Bar{\calX}$, the 2-D IFFT of $\Bar{\calX_i}$ denoted by $\Bar{X_i}$ is expressed as:
\begin{equation} \label{eq:IFFT}
    \Bar{X_i}(k, l) = \frac{1}{HW}\sum_{u=0}^{H-1}\sum_{v=0}^{W-1} \Bar{\calX_i}(u, v)e^{{i2\pi} (\frac{uk}{H} + \frac{vl}{W})}.
\end{equation}
Formally, let $g(\calX, K)$ be the output of the global filtering as above, $h$ be the high pass filter, $\mathbb{F}$ and $\mathbb{F}^{-1}$ be the FFT and the inverse IFFT, respectively, the output of the global branch is calculated as:
\begin{equation}
    \calF_{global}(X) = \mathbb{F}^{-1}(h(g(\mathbb{F}(X), K)),
\end{equation} 
where $\mathbb{F}, h, \mathbb{F}^{-1}$ are applied in a channel-wise fashion. 

The FAM module also consists of a local branch, which is a $1\times 1$ convolutional layer in the spatial domain. This layer aims to leverage the information of features in the spatial domain. Let $\calF_{local}(X)$ be the output of the local branch, and the output of the frequency attention module is calculated as below:
\begin{equation}
    \calF_{out} = \gamma_1 * \calF_{global} + \gamma_2 * \calF_{local},
    \label{eq:F_out}
\end{equation}
where $\gamma_1$ and $\gamma_2$ are the learnable weighting parameters of the global and local branches, respectively. 
\paragraph{Computational complexity of the FAM module.} 
The global branch comprises a fast Fourier transform (FFT), an inverse fast Fourier transform (IFFT), a global filter, and a high pass filter (HPF).
 
The complexity of the FFT of an image with dimensions $H \times W$ is $\mathcal{O}(HW log(HW))$. Similarly, the complexity of the inverse fast Fourier transform (IFFT) of a frequency image with dimensions $H\times W$ is $\mathcal{O}(HW log(HW))$.
Therefore, the complexities of the FFT, global filter, HPF, and IFFT components in the FAM module are $\mathcal{O}(C_{in} HWlog (HW))$,  $\mathcal{O}(C_{out}C_{in}HW)$, $\mathcal{O}(C_{out} HW)$,  and $\mathcal{O}(C_{out} HWlog (HW))$, respectively. 

The FAM module also consists of a local branch, which is a 1 × 1 convolutional layer in the spatial domain. 
The local branch has the complexity of $\mathcal{O}(C_{out}C_{in}HW)$.
Overall, the FAM module has the complexity of $\mathcal{O}(C_{out}C_{in}HW)$.

 \vspace{0.2cm}
\subsection{Applying FAM to knowledge distillation}
\vspace{0.2cm}
\subsubsection{Layer-to-layer intermediate feature-based knowledge distillation}
\label{subsec:layer-to-layer}
\begin{figure}[]
\centering
  \includegraphics[scale=1, width=1\linewidth]{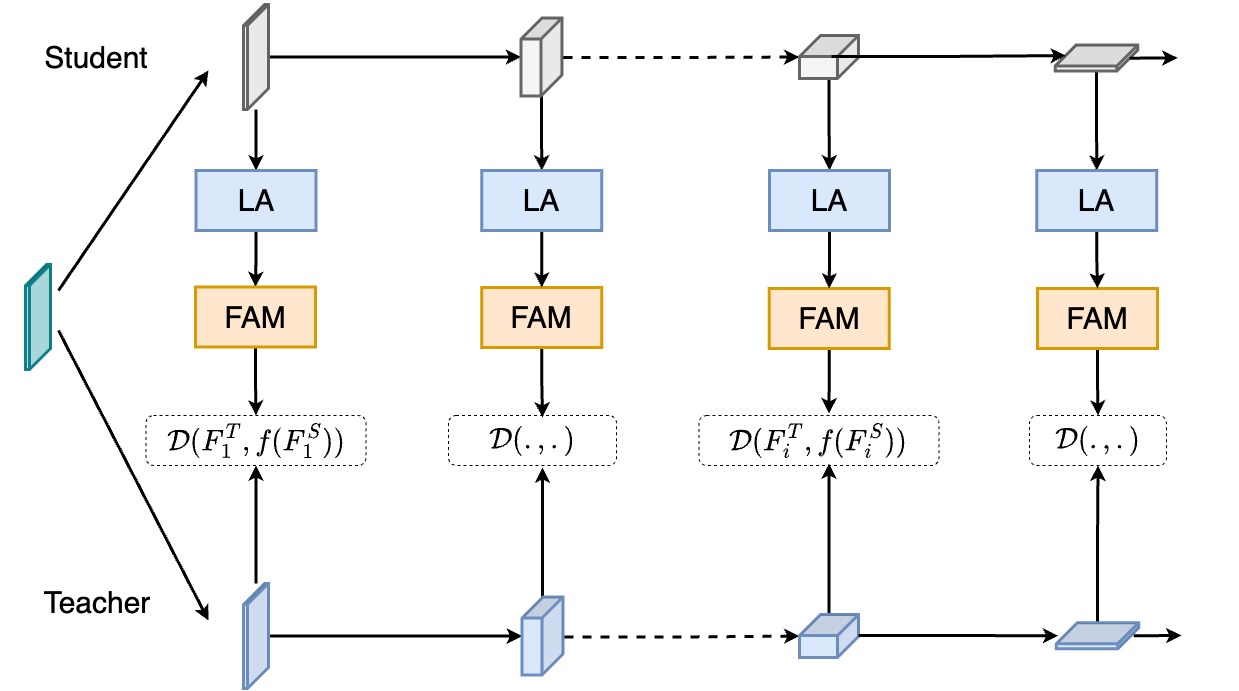}
    \caption{The proposed enhanced layer-to-layer knowledge distillation. LA is the  local attention and FAM is the proposed frequency attention module. $\calD$ is the distance function. $F^T$ and $F^S$ represent the feature maps of teacher and student, respectively.}
\label{fig:one-to-one}       
\end{figure}

\begin{figure}[]
\centering
  \includegraphics[scale=1, width=1\linewidth]{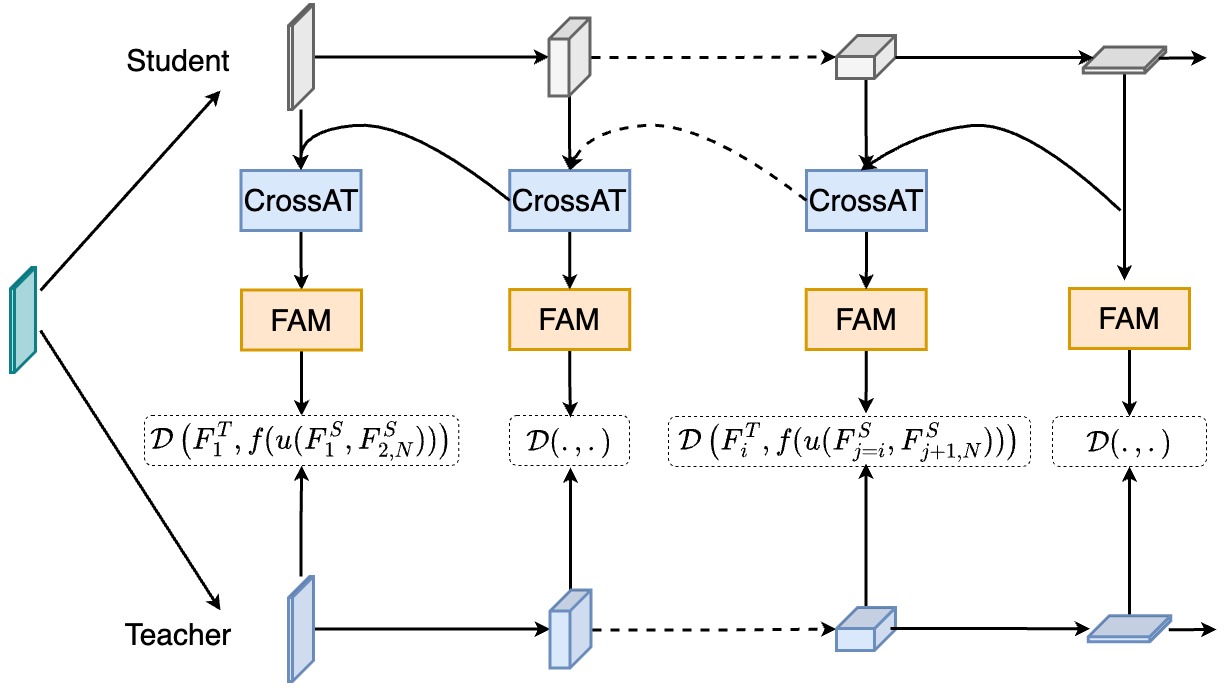}
    \caption{The proposed enhanced knowledge review distillation. CrossAT is the cross attention and FAM is the proposed frequency attention module. $\calD$ is the distance function. $F^T$ and $F^S$ represent the feature maps of teacher and student, respectively.}
\label{fig:one-to-many}       
\end{figure}
Let $\calI$ be the selected layer indices from the teacher for intermediate feature-based distillation. The layer-to-layer knowledge distillation loss is defined as
\begin{equation}
    \calL_{feat} = \sum_{i \in \calI} \calD \left(F_{i}^{T}, f(F_{j}^{S})\right),
    \label{eq:fea_loss}
\end{equation}
where $F_{j}^{S}$ is the feature map from the $j^{th}$ layer of the student selected for receiving the knowledge from the feature map  $F_{i}^{T}$ from $i^{th}$ layer of the teacher; $f$ is a transformation applied on the student's feature map. In our work, $f$ is the FAM module. $\calD$ is a distance function. In this work, we use $L_2$ distance as the distance function. It is worth noting the teacher is fixed in our framework, i.e., there are no transformations applied on teacher's feature maps.

\vspace{0,3cm}
In order to make FAM to better mimic teacher, we find that it would be beneficial to also enhance local structures in the spatial domain. To this end, we place an attention layer after student's feature maps before feeding it through the FAM module, as shown in Figure~\ref{fig:one-to-one}. To avoid increasing model complexity, we use local self-attention (LA) layer introduced by \cite{Stand-alone-SA}. In LA, the self-attention is applied only to a small neighbourhood around each position. 

\subsubsection{Knowledge review distillation }
\label{subsec:reviewkd}

We {also} integrate the FAM module into the knowledge review distillation mechanism~\cite{ReviewKD}, as shown in Figure~\ref{fig:one-to-many}. In~\cite{ReviewKD}, the authors propose a knowledge review mechanism that uses teacher’s low-level features to supervise deeper student's features. 
They fuse different levels of the student's features before mimicking knowledge from the teacher. In knowledge review mechanism~\cite{ReviewKD}, the distillation loss is defined as follows:
\begin{equation} 
\small
    \calL_{feat} = \calD (F_{M}^{T}, f(F_{N}^{S})) +\sum_{i=M-1}^{1} \calD \left(F_{i}^{T}, f(u(F_{j=i}^{S}, F_{j+1,N}^S))\right),
  \label{eq:fea_loss2}
\end{equation}
where $M$ and $N$ are the numbers of selected intermediate layers of teacher and student used for knowledge distillation. We note that in intermediate feature-based KD, the student and the teacher models are often divided into stages. The number of stages is the same for the teacher and the student, i.e., $M=N$. The last layers in each stage are used for distillation. $u(.,.)$ is a fusion function that recursively fuses student features. $F_{j+1,N}^S$ denotes the fusion of features from $F_{j+1}^S$ to $F_{N}^S$; $u(F_{j=i}^{S}, F_{j+1,N}^S) = u(F_{j=i}^{S}, u(F_{j+1}^S,F_{j+2,N}^S))$; $f$ is the FAM module. 

\vspace{0.3cm}
In~\cite{ReviewKD}, $u(.,.)$ is an attention-based fusion (ABF~\cite{ReviewKD}) function that learns two attention maps for two inputs and uses attention maps to aggregate two inputs. 
In this work, we propose using cross attention~\cite{SA} in which the low-level feature map is considered as the value and key and the high (fused) feature map is considered as the query when fusing student's feature maps at different levels. 
Specifically, let $F' = u(F_{j+1}^S,F_{j+2,N}^S))$
\begin{equation}\label{eqn:crossat}
    u(F_{j}^S, F') = softmax((\bW_QF')(\bW_KF_j^S)^T)\bW_VF_j^S,
\end{equation}
where $\bW_Q$, $\bW_K$, and $\bW_V$ represent learnable parameters for query, key, and value, respectively.
In summary, compared to~\cite{ReviewKD}, firstly, to emphasize the importance of the student’s feature map that is at the same level as the teacher’s feature map, our enhanced KD review architecture uses cross attention instead of ABF~\cite{ReviewKD}. Then, we feed the output of cross attention to the FAM module to adjust frequencies before computing the distance function $\calD$. 
 
The overall loss consists of the task loss (i.e., cross-entropy loss for classification task) and the feature distillation loss:
\begin{equation}\label{eqn:final_loss}
    \calL = \calL_{task} + \alpha \calL_{feat} 
\end{equation}

\section{Experiments}
\label{sec:experiments}
\subsection{Experimental setup}
\label{subsec:implementations}
\paragraph{Datasets.}
We evaluate our approach on  {CIFAR-100} \cite{cifar} and {ImageNet} \cite{imagenet} datasets for image classification task, and COCO dataset \cite{COCO} for object detection task. The CIFAR-100 dataset consists of $60,000$ images for $100$ classes, in which, $50,000$ and $10,000$ images are used for training and validation sets, respectively. ImageNet is {a} challenging dataset with $1000$ classes. 
This dataset contains $1.2$ million images for training and $50,000$ images for validation, which is used as a test set in our experiments. For object detection task, COCO is a standard dataset with multiple objects in an image. In total, this dataset contains $1.5$ million object instances of $80$ object categories in $118,000$ training and $5,000$ validation images.
\paragraph{Implementation details.} 
We apply our method across various teacher-student architecture pairs, as shown in Table~\ref{tab:sota_cifar}, Table~\ref{tab:cifar_dif_arch}, Table~\ref{tab:sota_imagenet}, and Table~\ref{tab:MS-COCO}. For a fair comparison, we do experiments on standard teacher/student pairs following other distillation methods \cite{CRD, WCoRD, ReviewKD, DKD} and {base on} the public distiller code-base \cite{DKD}. This includes the distillation when teachers and students are in the same architecture and in different architectures. For training, we use the standard training procedure following \cite{ReviewKD, DKD} and pre-trained teachers in all settings for both classification and object detection tasks. We employ $L_2$ distance as a distance function $\calD$ when calculating the $\calL_{feat}$ losses (Eq.~(\ref{eq:fea_loss}) and Eq.~(\ref{eq:fea_loss2})). 
The implementation details for CIFAR-100, ImageNet, and MS-COCO datasets and the values of the hyper-parameter $\alpha$ (Eq.~\ref{eqn:final_loss}) for each teacher/student pair are provided in supplementary materials due to page limit.  
\begin{table*}[t]
\def\arraystretch{1.05}
\centering
 \begin{tabular}{c|c|c|c|c|c}
    \hline
   Teacher & WRN-40-2 &  WRN-40-2 & ResNet56 & ResNet110 & ResNet32x4 \\
    Student & WRN-16-2 &  WRN-40-1 & ResNet20 & ResNet32 & ResNet8x4 \\
    \hline
    Teacher & 75.61 & 75.61 & 72.34 & 74.31 & 79.42 \\
    Student & 73.26 & 71.98 & 69.06 & 71.14 & 72.50 \\
    \hline
    \multicolumn{6}{c}{ Soft logit-based distillation} \\
\hline
    KD \cite{KD} & 74.92 & 73.54 & 70.66 & 73.08 & 73.33 \\
    
     DKD \cite{DKD} & {76.24} & 74.81 & {71.97} & {74.11} & {76.32} \\
     \hline
    \multicolumn{6}{c}{ Layer to layer-based distillation} \\
    \hline
    FITNET \cite{FitNets} & 73.58 & 72.24 & 69.21 & 71.06 & 73.50 \\
    AT \cite{AT}& 74.08 & 72.77 & 70.55 & 72.31 & 73.44 \\
    VID \cite{VID} & 74.11 & 73.30 & 70.38 & 72.61 & 73.09 \\
    RKD \cite{RKD} & 73.35 & 72.22 & 69.61 & 71.82 & 71.90 \\
    CRD \cite{CRD} & 75.48 & 74.14 & 71.16 & 73.48 & 75.51 \\
    
    WCoRD \cite{WCoRD}& 75.88 & 74.73 & 71.56  & 73.81 & 75.95 \\
    
    
    OFD \cite{OFD}& 75.24 & 74.33 & 70.98  & 73.23 & 74.95 \\
    

    FAM-KD (layer-to-layer) - Ours & 76.03 & 74.88 & 72.03  & 74.03 & 76.24 \\
     \hline
     \multicolumn{6}{c}{Layer to layer + Soft logit-based distillation} \\
     \hline
    WCoRD + KD\cite{WCoRD}& 76.11 & 74.72 & 71.92  & 74.20 & 76.15 \\
    
    \hline
        \multicolumn{6}{c}{ Knowledge review-based distillation} \\
    \hline
    ReviewKD \cite{ReviewKD}& 76.12 & {75.09} & 71.89 & 73.89 & 75.63 \\

    FAM-KD (review) - Ours & \textbf{76.47} & \textbf{75.40} & \textbf{72.15} & \textbf{74.45} & \textbf{76.84}\\
    
   \hline

    \hline
 \end{tabular}
\vspace{-0.16cm}
\caption{Results on the CIFAR-100 validation set. Teachers and students are in the same architecture. 
FAM-KD (layer-to-layer) and FAM-KD (review) refer to our proposed methods in Section~\ref{subsec:layer-to-layer} and Section~\ref{subsec:reviewkd}, respectively. Our reported results are an average of three trials. 
} 
\label{tab:sota_cifar} 
\end{table*}
\begin{table*}[!t]
\def\arraystretch{1.05}
\centering
 \begin{tabular}{c|c|c|c|c}
    \hline
   Teacher & ResNet32x4 &  WRN-40-2 & ResNet32x4 & VGG13   \\
    Student & ShuffleNet-V1 &  ShuffleNet-V1 & ShuffleNet-V2 & MobileNet-V2  \\
    \hline
    Teacher & 79.42 & 75.61 & 79.42 & 74.64   \\
    Student & 70.50 & 70.50 & 71.82 & 64.60   \\
    \hline
        \multicolumn{5}{c}{Soft logit-based distillation} \\
        \hline
    KD \cite{KD} & 74.07 & 74.83 & 74.45 & 67.37   \\
     DKD \cite{DKD} & 76.45 & 76.70 &  77.07 & 69.71 \\
    \hline
            \multicolumn{5}{c}{Layer to layer-based distillation} \\
        \hline
    FITNET \cite{FitNets} & 73.59 & 73.73 & 73.54 & 63.16  \\
    AT \cite{AT}& 71.73 & 73.32 & 72.73 & 59.40   \\
    VID \cite{VID} & 73.38 & 73.61 & 73.40 & 65.56   \\
    
    RKD \cite{RKD} & 72.28 & 72.21 & 73.21 & 64.52   \\
    CRD \cite{CRD} & 75.11 & 76.05  & 75.65 & 69.73  \\
    WCoRD \cite{WCoRD}& 75.40 & 76.32 & 75.96 & 69.47     \\

    OFD \cite{OFD}& 75.98 & 75.85 &  76.82 & 69.48   \\

   FAM-KD (layer-to-layer) - Ours & 77.15 & 77.33 & 77.64 & 69.96    \\
      \hline
           \multicolumn{5}{c}{Layer to layer + Soft logit based-distillation} \\
           \hline
               WCoRD + KD \cite{WCoRD}& 75.77 & 76.68 & 76.48 & 70.02     \\
        \hline
           \multicolumn{5}{c}{Knowledge review-based distillation} \\
        \hline
        
       ReviewKD \cite{ReviewKD}& 77.45 & 77.14 & 77.78 & 70.37   \\

    FAM-KD (review) - Ours & \textbf{77.76} & \textbf{77.57} & \textbf{78.41} & \textbf{70.88}   \\
    \hline
 \end{tabular}
\vspace{-0.16cm}
\caption{The comparative results on the CIFAR-100 validation set. Teachers and students are in the different architectures. FAM-KD (layer-to-layer) and FAM-KD (review) refer to our proposed methods in Section~\ref{subsec:layer-to-layer} and Section~\ref{subsec:reviewkd}, respectively. 
Our reported results are an average of three trials.
} 
\label{tab:cifar_dif_arch} 
\end{table*}
\begin{table*}[!htbp]
\centering
\resizebox{2.0\columnwidth}{!}{
 \begin{tabular}{c|ccccccccccc}
    \hline
     Setting & &Teacher &  Student & KD \cite{KD} & AT \cite{AT} & OFD \cite{OFD}&
     CRD \cite{CRD} &
     WCoRD \cite{WCoRD} &  DKD \cite{DKD} & ReviewKD \cite{ReviewKD} & FAM-KD (Ours)\\
     \hline
    \multirow{2}{*}{(a)} & Top-1 & 73.31 & 69.75 & 70.66 & 70.69 & 70.81 & 71.17 & 71.49 & 71.70 & 71.61  & \textbf{71.91} \\
    
     & Top-5 & 91.42 & 89.07 & 89.88 & 90.01 & 89.98 & 90.13 & 90.16 & 90.41 & {90.51}  & \textbf{90.53}  \\
     \hline

     \multirow{2}{*}{(b)} & Top-1 & 76.16 & 68.87 & 68.58 & 70.69 & 70.81 & 71.17 & - & 72.05 & 72.56  & \textbf{73.33} \\
    
     & Top-5 & 92.86 & 88.76 & 88.98 & 90.01 & 89.98 & 90.13 & - & 91.05 & {91.00}  & \textbf{91.44}  \\
     \hline
\end{tabular}}
\caption{Top-1 and top-5 accuracy (\%) on the ImageNet validation set. (a) ResNet34 and ResNet18 and (b) ResNet50 and MobileNetV1 are used as the teacher and student architectures. Our results (FAM-KD) are with the enhanced knowledge review-based distillation (Section~\ref{subsec:reviewkd}). Our reported results are an average of three trials.
} 
\label{tab:sota_imagenet} 
\end{table*}
\begin{figure*}[t]
\centering
      \begin{subfigure}[b]{0.195\textwidth}
         \centering
         \includegraphics[width=\textwidth]{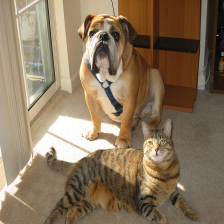}
         \caption{Original image}
     \end{subfigure}
     \hfill
    \begin{subfigure}[b]{0.195\textwidth}
         \centering
         \includegraphics[width=\textwidth]{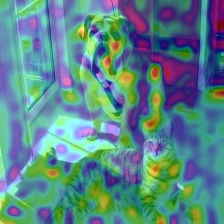}
         \caption{ResNet18 (w/o KD)}
     \end{subfigure}
    \hfill
    \begin{subfigure}[b]{0.195\textwidth}
         \centering
         \includegraphics[width=\textwidth]{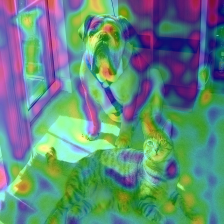}
         \caption{OFD~\cite{OFD}}
     \end{subfigure}
     \hfill
     \begin{subfigure}[b]{0.195\textwidth}
         \centering
         \includegraphics[width=\textwidth]{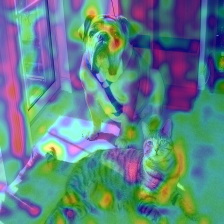}
         \caption{Knowledge review~\cite{ReviewKD}}
     \end{subfigure}
          \hfill
    \begin{subfigure}[b]{0.195\textwidth}
    \centering
         \includegraphics[width=\textwidth]{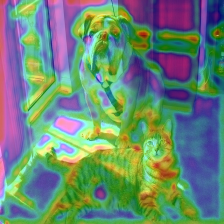}
         \caption{FAM-KD (Ours)}
     \end{subfigure}
    \caption{(a) Original image. (b) - (e) Grad-CAMs~\cite{Grad-cam} from layer 9 of ResNet18 model when training (b) without knowledge distillation, (c) with OFD~\cite{OFD}, (d) with knowledge review~\cite{ReviewKD}, and (e) with FAM-KD (ours), respectively. When training with distillation, ResNet34 is used as the teacher. The figure shows that our FAM-KD (e) has better focus on the object than using OFD~\cite{OFD} and knowledge review~\cite{ReviewKD}.}
    \label{fig:visualization}       
\end{figure*}

\begin{center}
\begin{table*}[t]
\centering{}%
\resizebox{1.0\columnwidth}{!}{
\begin{tabular}{c|ccc|ccc}
\hline
\multirow{2}{*}{Method} & \multicolumn{3}{c}{ResNet101 \& ResNet18} & \multicolumn{3}{|c}{ResNet101 \& ResNet50} \tabularnewline
 & AP & AP\_50 & AP\_75 & AP & AP\_50 & AP\_75  \\ 
\hline 
Teacher & 42.04  & 62.48  & 45.88  & 42.04  & 62.48  & 45.88  \\ 
Student & 33.26  & 53.61  & 35.26  & 37.93  & 58.84  & 41.05  \\ 
\hline
KD \cite{KD} & 33.97  & 54.66  & 36.62  & 38.35  & 59.41  & 41.71 \tabularnewline
FitNet \cite{FitNets} & 34.13  & 54.16  & 36.71  & 38.76  & 59.62  & 41.80 \tabularnewline
FGFI \cite{FGFI} & 35.44  & 55.51  & 38.17  & 39.44  & 60.27  & 43.04 \tabularnewline
ReviewKD \cite{ReviewKD} & 36.75  & 56.72  & 34.00  & 40.36  & 60.97  & 44.08 \tabularnewline
DKD \cite{DKD}& 35.05  & 56.60  & 37.54  & 39.25  & 60.90  & 42.73 \tabularnewline
DKD + ReviewKD \cite{DKD} & 37.01  & 57.53  & 39.85  & 40.65  & \textbf{61.51}  & 44.44 \tabularnewline
\hline
FAM-KD (ours) & \textbf{37.20} & \textbf{57.86} & \textbf{40.01}&	\textbf{40.77}&	61.42 &	\textbf{44.49} \tabularnewline
\hline
\end{tabular}}
\caption{Comparative object detection accuracy on the MS-COCO dataset. We use the two-stage method Faster RCNN \cite{FasterRCNN} with FPN \cite{FPN} as the detector. On the student side, ResNet18  and ResNet50  models are selected as backbones, while teacher models use ResNet101  as a backbone. Our results (FAM-KD) are with the enhanced knowledge review-based distillation (Section~\ref{subsec:reviewkd}). Our reported results are an average of three trials. 
}
\label{tab:MS-COCO}
\end{table*}
\par\end{center}

\subsection{Comparison with the state of the art}
\subsubsection{Image classification}
\paragraph{Comparative results on CIFAR-100.} We present top-1 classification accuracy on the CIFAR-100 by various teacher-student pairs, both from the same network family (Table \ref{tab:sota_cifar}) and from the different network family (Table \ref{tab:cifar_dif_arch}). The selected networks comprise ResNet~\cite{resnet}, WideResNet~\cite{WRN}, ShuffleNet~\cite{ShuffleNet}, MobileNetV2 \cite{Mobilenetv2}, and VGG \cite{VGG}. The results of competitors are cited from~\cite{WCoRD, DKD, ReviewKD}. 

Overall, our method FAM-KD (review) consistently outperforms all compared methods in all settings. 
In some cases, i.e., WRN-40-2/WRN-16-2, ResNet110/ResNet32, WRN-40-2/ShuffleNet-V1, students' performance even surpasses the teachers. 

\vspace{0.3cm}
Regarding layer-to-layer distillation, our method FAM-KD (layer-to-layer) outperforms all other methods belonging to the same category. 
Our method consistently outperforms the most competitor WCoRD~\cite{WCoRD} on all settings. 
Compare to WCoRD + KD~\cite{WCoRD}, our method achieves competitive results, despite that we only use feature-based distillation. The highest improvement over WCoRD + KD~\cite{WCoRD} is $1.38\%$ with the ResNet32x4/ShuffleNet-V1 setting.  
It is worth noting that even with the layer-to-layer setting, the FAM-KD achieves comparable results with the current state-of-the-art feature distillation method using the review mechanism \cite{ReviewKD}. 

Regarding the knowledge review mechanism (FAM-KD (review)), we outperform compared methods for all teacher-student distillation pairs. 
Compare to \cite{ReviewKD}, our method outperforms ReviewKD \cite{ReviewKD} in all {cases}. 
Compare to DKD~\cite{DKD}, which is a soft logit-based distillation method, our method FAM-KD (review) also outperforms DKD~\cite{DKD} in all settings.  The highest improvement is $1.34\%$ with the ResNet32x4/ShuffleNet-V2 setting. The promising results have shown the effectiveness of the FAM module, supporting students to perform better. 
\paragraph{Comparative results on ImageNet.} 
We validate our approach on the large-scale dataset ImageNet \cite{imagenet} in the case of integrating the FAM module into the knowledge review distillation mechanism (FAM-KD). Table~\ref{tab:sota_imagenet} presents the top-1 and top-5 classification accuracy on the ImageNet validation set of various distillation methods. When both teacher and student have the same architecture, we employ ResNet34/ResNet18 as the teacher/student pair. Meanwhile, when teacher and student have different architectures, we use ResNet50/MobileNetV1 as the teacher/student pair. Our approach yields the highest performance on both top-1 and top-5 accuracy.
For ResNet34/ResNet18, compared to the vanilla KD \cite{KD}, the FAM-KD improves by a large margin of $1.24\%$ top-1 accuracy. Meanwhile, compared to  ReviewKD \cite{ReviewKD} and DKD \cite{DKD}, the FAM-KD improves $0.3\%$ and $0.21\%$ top-1 accuracy, respectively. The relative improvements\footnote{\label{note1}Similar to \cite{WCoRD}, we compute the relative improvement as $\frac{\mathrm{Ours} - \mathrm{A}}{\mathrm{A} - \mathrm{KD}}$, 
where $\mathrm{A}$ is the method we are comparing to. For each method, the corresponding accuracy of the student is used for the calculation.} over DKD and ReviewKD are considerable at 20.2\% and 31.6\%, respectively. For ResNet50/MobileNetV1, our approach yields a significant improvement, i.e., the improvements of 1.28\% and 0.77\% over DKD and ReviewKD in top-1 accuracy, and the corresponding relative improvements over DKD and ReviewKD are 36.9\% and 19.3\%, respectively.

Figure~\ref{fig:visualization} visualizes Grad-CAMs ~\cite{Grad-cam} (gradient-weighted class activation mapping) extracted from layer 9 of the ResNet18, \blue{which serves as student's architecture, with ResNet34 used as the teacher}. The figure shows that the Grad-CAM when using our FAM-KD (e) has better focus on the object than using OFD~\cite{OFD} and knowledge review~\cite{ReviewKD}. 

\subsubsection{Object detection}
Table~\ref{tab:MS-COCO} presents the object detection accuracy on the MS COCO dataset. We use the FasterRCNN~\cite{FasterRCNN} with FPN~\cite{FPN} as the detector, and use teacher/student pairs ResNet101/ResNet18, ResNet101/ResNet50 for the backbones. The results show that our method FAM-KD consistently outperforms ReviewKD~\cite{ReviewKD} and the recent work DKD~\cite{DKD} for both settings at all metrics. With the teacher/student pairs ResNet101/ResNet18 and ResNet101/ResNet50, the proposed method outperforms DKD $2.15$ and $1.52$ AP points, respectively.
In~\cite{DKD}, to boost the detection accuracy of the student, the authors combine their soft logit-based distillation DKD with the knowledge review-based distillation~\cite{ReviewKD}. Compare to DKD+ReviewKD~\cite{DKD}, despite that we only use intermediate feature-based distillation, our method outperforms DKD+ReviewKD for most metrics, except AP\_50 with ResNet101/ResNet50 setting. 
\vspace{0.3cm}
\subsection{Ablation studies}
\label{subsec:ablation}
\begin{center}
\begin{table}[t]
\centering{}%
\begin{tabular}{c|cc|c}
\hline
Setting & Global branch & Local branch & Top-1\tabularnewline
\hline
(a) &  &   \checkmark & 73.90\tabularnewline
(b) & \checkmark &  & 74.17\tabularnewline
(c) & \checkmark  & \checkmark & 74.45\tabularnewline
\hline
\end{tabular}
\caption{Impact of the global branch and local branch of the FAM.
ResNet110 and ResNet32 are used as the teacher and the student, respectively. The results are on the CIFAR-100 validation set.} 
\label{tab:ab_global_local}
\end{table}
\par\end{center}
In this section, we focus on investigating how different components of the FAM module contribute to the performance of FAM-KD. All experiments are conducted on CIFAR-100 dataset with ResNet110 as a teacher and ResNet32 as a student when integrating the FAM module into the knowledge review-based mechanism as presented in Section ~\ref{subsec:reviewkd}. Other experiments that inspect the effectiveness of the FAM module when integrating the FAM module to ReviewKD \cite{ReviewKD}, given the same ReviewKD implementation in the public distiller code-base \cite{DKD} are provided in supplementary materials. 

\paragraph{With/without global and local branches in FAM.} In Table \ref{tab:ab_global_local}, we present performance of FAM-KD with and without the global branch and local branch. The results show that the global branch benefits the model better than the local one and having both branches gives the best result.
\begin{center}
\begin{table}[t]
\centering{}%
\begin{tabular}{c|c}
\hline
Setting & Top-1 \\
\hline
FAM-KD (w/o HPF)  & 73.82 \tabularnewline
FAM-KD  & 74.45 \tabularnewline
\hline
\end{tabular}
\caption{Effect of the high pass filter (HPF) component in global branch of the FAM. ResNet110 and ResNet32 are used as the teacher and the student, respectively. The results are on the CIFAR-100 validation set.}
\label{tab:ab_hpf}
\end{table}
\par\end{center}
\paragraph{With/without high pass filter (HPF) in the global branch.} The effectiveness of the HPF is presented in Table~\ref{tab:ab_hpf}, i.e., having the HPF boosts the performance $0.63\%$. This shows the effectiveness of HPF, which helps to filter out the lowest frequency components and encourages the student to de-focus from the non-salient regions.

\section{Conclusion}
\label{sec:conclusion}
In this paper, we propose
to use the frequency domain to encourage the student model to capture
both detailed and higher-level information such as object
parts based on a well-trained teacher's guidance. We introduce a novel frequency attention module (FAM) for knowledge distillation that operates in the frequency domain and has a filter that can be adjusted to mimic the teacher's features.
This encourages the student's features to have similar geometric structures to the teacher's features.
Moreover, we propose an enhanced knowledge review-based distillation by leveraging the proposed FAM and cross attention. We extensively evaluate our approach with different teacher and student models, and the proposed approach achieves significant improvements compared to other state-of-the-art methods for image classification on the CIFAR-100 and ImageNet datasets and for object detection on the MS COCO dataset.

{\small
\bibliographystyle{ieee_fullname}
\bibliography{ref}
}

\end{document}